\documentclass[11pt,a4paper]{article}
\usepackage[hyperref]{acl2021}
\usepackage{times}
\usepackage{latexsym}

\usepackage{microtype}
\usepackage{graphicx}
\usepackage{subfigure}
\usepackage{booktabs} 
\usepackage{url}
\usepackage{times}
\usepackage{latexsym}
\usepackage{array}
\usepackage{adjustbox}
\usepackage{multirow}
\usepackage{hyperref}
\usepackage{longtable}

\newcommand{\insertXNLI}{
    \begin{table*}[ht]
        \begin{center}
            \resizebox{1\linewidth}{!}{
            \begin{tabular}[b]{l|c| ccccccccccccccc c}
            \toprule
            {\bf Model} & {\bf Data (\#tok)} & {\bf en} & {\bf fr} & {\bf es} & {\bf de} & {\bf el} & {\bf bg} & {\bf ru} & {\bf tr} & {\bf ar} & {\bf vi} & {\bf th} & {\bf zh} & {\bf hi} & {\bf sw} & {\bf ur} & {\bf Avg}\\
                \midrule
                
                \multicolumn{17}{l}{\it Fine-tune multilingual model on English training set (Cross-lingual Transfer)} \\
                \midrule
                mBERT &  \multirow{2}{*}{Wikipedia} & 80.8 & 64.3 & 68.0 & 70.0 & 65.3 & 73.5 & 73.4 & 58.9 & 67.8 & 49.7 & 54.1 & 60.9 & 57.2 & 69.3 & 67.8 & 65.4 \\
                XLM & & 83.2 & 76.5 & 76.3 & 74.2 & 73.1 & 74.0 & 73.1 & 67.8 & 68.5 & 71.2 & 69.2 & 71.9 & 65.7 & 64.6 & 63.4 & 71.5 \\
                \midrule
                mT5-Base & \multirow{3}{*}{mC4} & 84.7 & 73.3 & 78.6 & 77.4 & 77.1 & 80.3 & 79.1 & 70.8 & 77.1 & 69.4 & 73.2 & 72.8 & 68.3 & 74.2 & 74.1 & 75.4 \\
                mT5-Large & & 89.4 & 79.8 & 84.1 & 83.4 & 83.2 & 84.2 & 84.1 & 77.6 & 81.5 & 75.4 & 79.4 & 80.1 & 73.5 & 81.0 & 80.3 & 81.1 \\
                mT5-XL & \multirow{1}{*}{(6.4T)} & 90.6 & 82.2 & 85.4 & 85.8 & 85.4 & 81.3 & 85.3 & 80.4 & 83.7 & 78.6 & 80.9 & 82.0 & 77.0 & 81.8 & 82.7 & 82.9 \\
                mT5-XXL & & 91.6 & 84.5 & 87.7 & 87.3 & 87.3 & 87.8 & 86.9 & 83.2 & 85.1 & 80.3 & 81.7 & 83.8 & 79.8 & 84.6 & 83.6 & 84.5 \\
                \midrule
                XLM-R\textsubscript{Base} & \multirow{3}{*}{CC100} & 85.8 & 79.7 & 80.7 & 78.7 & 77.5 & 79.6 & 78.1 & 74.2 & 73.8 & 76.5 & 74.6 & 76.7 & 72.4 & 66.5 & 68.3 & 76.2 \\
                \xlml & & 89.1 & 84.1 & 85.1 & 83.9 & 82.9 & 84.0 & 81.2 & 79.6 & 79.8 & 80.8 & 78.1 & 80.2 & 76.9 & 73.9 & 73.8 & 80.9 \\
                \xlmxl & \multirow{1}{*}{(167B)} & 90.7 & 85.5 & 86.5 & 84.6 & 84.0 & 85.2 & 82.7 & 81.7 & 81.6 & 82.4 & 79.4 & 81.7 & 78.5 & 75.3 & 74.3 & 82.3 \\
                \xlmxxl & & 91.6 & 86.2 & 87.3 & 87.0 & 85.1 & 85.7 & 82.5 & 82.0 & 82.5 & 83.0 & 79.5 & 82.6 & 79.8 & 76.2 & 74.9 & 83.1 \\
                \midrule
                \multicolumn{17}{l}{\it Translate everything to English and use English-only model (TRANSLATE-TEST)} \\
                \midrule
                RoBERTa & CC-En & 91.3 & 82.9 & 84.3 & 81.2 & 81.7 & 83.1 & 78.3 & 76.8 & 76.6 & 74.2 & 74.1 & 77.5 & 70.9 & 66.7 & 66.8 & 77.8 \\
                \midrule
                \multicolumn{17}{l}{\it Fine-tune multilingual model on all training sets (TRANSLATE-TRAIN-ALL)} \\
                \midrule
                mT5-Base & \multirow{3}{*}{mC4} & 82.0 & 74.4 & 78.5 & 77.7 & 78.1 & 79.1 & 77.9 & 72.2 & 76.5 & 71.5 & 75.0 & 74.8 & 70.4 & 74.5 & 76.0 & 75.9 \\
                mT5-Large & & 88.3 & 80.3 & 84.1 & 84.0 & 83.7 & 84.9 & 83.8 & 79.8 & 82.0 & 76.4 & 79.9 & 81.0 & 75.9 & 81.3 & 81.7 & 81.8 \\
                mT5-XL & \multirow{1}{*}{(6.4T)} & 90.9 & 84.2 & 86.8 & 86.8 & 86.4 & 87.4 & 86.8 & 83.1 & 84.9 & 81.3 & 82.3 & 84.4 & 79.4 & 83.9 & 84.0 & 84.8 \\
                mT5-XXL & & \bf92.7 & 87.2 & \bf 89.4 & \bf 89.8 & \bf 89.5 & \bf 90.0 & \bf 89.1 & \bf 86.5 & \bf 87.6 & 84.3 & \bf 85.6 & \bf 87.1 & 83.8 & \bf 87.5 & \bf 86.5 & \bf 87.8 \\
                \midrule
                XLM-R\textsubscript{Base} & \multirow{3}{*}{CC100} & 85.4 & 81.4 & 82.2 & 80.3 & 80.4 & 81.3 & 79.7 & 78.6 & 77.3 & 79.7 & 77.9 & 80.2 & 76.1 & 73.1 & 73.0 & 79.1 \\
                \xlml & & 89.1 & 85.1 & 86.6 & 85.7 & 85.3 & 85.9 & 83.5 & 83.2 & 83.1 & 83.7 & 81.5 & 83.7 & 81.6 & 78.0 & 78.1 & 83.6 \\
                \xlmxl & \multirow{1}{*}{(167B)} & 91.1 & 87.2 & 88.1 & 87.0 & 87.4 & 87.8 & 85.3 & 85.2 & 85.3 & 86.2 & 83.8 & 85.3 & 83.1 & 79.8 & 78.2 & 85.4 \\
                \xlmxxl & & 91.5 & \bf 87.6 & 88.7 & 87.8 & 87.4 & 88.2 & 85.6 & 85.1 & 85.8 & \bf  86.3 & 83.9 & 85.6 & \bf 84.6 & 81.7 & 80.6 & 86.0 \\

                \bottomrule
            \end{tabular}
            }
            \caption{\textbf{Results on cross-lingual classification (XNLI).} We report the accuracy on each of the 15 XNLI languages and average accuracy, and specify the dataset and its corresponding size in number of tokens. We report results of XLM-R models with increasing capacity, from 270M (Base), 550M (Large), 3.5B (XL) to 10.7B (XXL) parameters. 
            \label{tab:xnli}}
        \end{center}
       \vspace{-0.6cm}
    \end{table*}
}

\newcommand{\insertMLQA}{
\begin{table*}[ht]
    \begin{center}
        \resizebox{0.8\linewidth}{!}{
        \begin{tabular}[ht]{l|cccccccc}
        \toprule
            {\bf Model} & {\bf en} & {\bf es} & {\bf de} & {\bf ar} & {\bf hi} & {\bf vi} & {\bf zh} & {\bf Avg} \\
            \midrule
            \multicolumn{9}{l}{\it Cross-lingual zero-shot transfer (models fine-tune on English data only)} \\
            \midrule
            mT5-Large & 84.9 / 70.7 & 65.3 / 44.6 & 68.9 / 51.8 & 73.5 / 54.1 & 66.9 / 47.7 & 72.5 / 50.7 & 66.2 / 42.0 & 71.2 / 51.7 \\
            mT5-XL & 85.5 / 71.9 & 68.0 / 47.4 & 70.5 / 54.4 & 75.2 / 56.3 & 70.5 / 51.0 & 74.2 / 52.8 & 70.5 / 47.2 & 73.5 / 54.4 \\
            mT5-XXL & \bf 86.7 / 73.5 & \bf 70.7 / 50.4 & \bf 74.0 / 57.8 & \bf 76.8 / 58.4 & \bf 75.6 / 57.3 & \bf 76.4 / 56.0 & {\bf 71.8} / 48.8 & \bf 76.0 / 57.4 \\
            \midrule
            \xlml & 80.6 / 67.8 & 74.1 / 56.0 & 68.5 / 53.6 & 63.1 / 43.5 & 69.2 / 51.6 & 71.3 / 50.9 & 68.0 / 45.4 & 70.7 / 52.7  \\
            \xlmxl & 85.1 / 72.6 & 66.7 / 46.2 & 70.5 / 55.5 & 74.3 / 56.9 & 72.2 / 54.7 & 74.4 / 52.9 & 70.9 / 48.5 & 73.4 / 55.3 \\
            \xlmxxl & 85.5 / 72.4 & 68.6 / 48.4 & 72.7 / {\bf 57.8} & 75.4 / 57.6 & 73.7 / 55.8 & 76.0 / 55.0 & 71.7 / {\bf 48.9} & 74.8 / 56.6 \\

            \bottomrule
        \end{tabular}
            }
            \caption{MLQA results (F1/EM) for each language.
            \label{tab:mlqa}}
    \end{center}
\end{table*}
}

\newcommand{\insertXQuad}{
\begin{table*}[ht]
    \begin{center}
        \resizebox{1\linewidth}{!}{
        \begin{tabular}[ht]{l|cccccccccccc}
        \toprule
            {\bf Model} & {\bf en} & {\bf ar} & {\bf de} & {\bf el} & {\bf es} & {\bf hi} & {\bf ru} & {\bf th} & {\bf tr} & {\bf vi} & {\bf zh} & {\bf avg} \\
            \midrule
            \multicolumn{13}{l}{\it Cross-lingual zero-shot transfer (models fine-tune on English data only)} \\
            \midrule
            mT5-Large & 88.4 / 77.3 &  75.2 / 56.7 &  80.0 / 62.9 &  77.5 / 57.6 &  81.8 / 64.2 &  73.4 / 56.6 &  74.7 / 56.9 &  73.4 / 62.0 &  76.5 / 56.3 &  79.4 / 60.3 &  75.9 / 65.5 &  77.8 / 61.5 \\
            mT5-XL & 88.8 / 78.1 & 77.4 / 60.8 & 80.4 / 63.5 & 80.4 / 61.2 & 82.7 / 64.5 & 76.1 / 60.3 & 76.2 / 58.8 & 74.2 / 62.5 & 77.7 / 58.4 & 80.5 / 60.8 & 80.5 / 71.0 & 79.5 / 63.6 \\
            mt5-XXL & \bf 90.9 / 80.1 & {\bf 80.3} / 62.6 & {\bf 83.1} / 65.5 & 83.3 / {\bf 65.5} & \bf 85.1 / 68.1 & \bf 81.7 / 65.9 & 79.3 / 63.6 & \bf 77.8 / 66.1 & {\bf 80.2} / 60.9 & \bf 83.1 / 63.6 & \bf 83.1 / 73.4 & \bf 82.5 / 66.8 \\
            \midrule
            \xlml & 86.5 / 75.7 & 68.6 / 49.0 & 80.4 / 63.4 & 79.8 / 61.7 & 82.0 / 63.9 & 76.7 / 59.7 & 80.1 / 64.3 & 74.2 / 62.8 & 75.9 / 59.3 & 79.1 / 59.0 & 59.3 / 50.0 & 76.6 / 60.8 \\
            \xlmxl & 89.5 / 79.0 & 78.4 / 61.6 & 81.3 / 64.1 & 82.3 / 63.9 & 84.6 / 66.2 & 78.8 / 63.2 & 81.5 / 65.0 & 76.0 / 65.5 & 73.9 / 57.9 & 81.7 / 61.8 & 72.3 / 66.1 & 80.0 / 64.9 \\
            \xlmxxl & 89.3 / 79.4 & 80.1 / {\bf 63.7} & 82.7 / {\bf 65.8} & \bf 83.4 / 65.5 & 83.8 / 66.0 & 80.7 / 65.4 & \bf 82.4 / 65.4 & 76.6 / 65.6 & 76.8 / {\bf 61.7} & 82.2 / 63.0 & 74.1 / 67.4 & 81.1 / 66.3 \\
            \bottomrule
        \end{tabular}
            }
            \caption{XQuad results (F1/EM) for each language.
            \label{tab:xquad}}
    \end{center}
\end{table*}
}

\newcommand{\insertGlue}{
    \begin{table}[b]
        \begin{center}
            \resizebox{1\linewidth}{!}{
            \begin{tabular}[b]{l|c|ccccc|c}
            \toprule
                {\bf Model} & {\bf \#lgs} & {\bf MNLI} & {\bf QNLI} & {\bf QQP} & {\bf SST} & {\bf MRPC} & {\bf Avg}\\
                \midrule
                RoBERTa$^{\dagger}$ & 1 & 90.2 & 94.7 & 92.2 & 96.4 & \bf 90.9  & 92.9 \\
                \xlml & 100 & 88.9 & 93.8 & 92.3 & 95.0 & 89.5 & 91.9 \\
                \xlmxl & 100 & 90.4 & 94.9 & 92.5 & 96.6 & 90.4 & 93.0 \\
                \xlmxxl & 100 & \bf 90.9 & \bf 95.0 & \bf 92.6 & \bf 96.7 & 90.7 & \bf 93.2 \\
                \bottomrule
            \end{tabular}
            }
            \caption{GLUE dev results
            \label{tab:glue}}
        \end{center}
       \vspace{-0.4cm}
    \end{table}
}

\newcommand{\insertcomparison}{
    \begin{table}[b]
        \begin{center}
            \resizebox{1\linewidth}{!}{
            \begin{tabular}[b]{l|c|c|c|c|c|c}
            \toprule
                \multirow{2}{*}{\bf Model} & {\bf Number of} & {\bf Dataset} & {\bf Dataset} & {\bf Number of} & {\bf Batch} & {\bf Sequence} \\
                 & {\bf parameters} & {\bf name} & {\bf size} & {\bf training tokens} & {\bf size} & {\bf length} \\
                \midrule
                \xlml & 550M & CC100 & 167B & 6T & 8192 & 512 \\
                \xlmxl & 3.5B & CC100 & 167B & 0.5T & 2048 & 512 \\
                \xlmxxl & 10.7B & CC100 & 167B & 0.5T & 2048 & 512 \\
                mt5-XL & 3.7B & mC4 & 6.4T & 1T & 1024 & 1024 \\
                mt5-XXL & 13B & mC4 & 6.4T & 1T & 1024 & 1024 \\
                \bottomrule
            \end{tabular}
            }
            \caption{Comparison of datasets and pretraining details between XLM-R and mT5. We report dataset sizes and number of updates in terms of number of tokens.
            \label{tab:comp}}
        \end{center}
       \vspace{-0.4cm}
    \end{table}
}

\aclfinalcopy 
\def\aclpaperid{12} 

\usepackage{xspace}

\newcommand{\xlml}{XLM-R\textsubscript{Large}\xspace}
\newcommand{\xlmxl}{XLM-R\textsubscript{XL}\xspace}
\newcommand{\xlmxxl}{XLM-R\textsubscript{XXL}\xspace}
\newcommand{\mbert}{mBERT\xspace}

\title{Larger-Scale Transformers for Multilingual Masked Language Modeling}

\author{Naman Goyal \space\space\space Jingfei Du \space\space\space Myle Ott \space\space\space Giri Anantharaman \space\space\space Alexis Conneau
\\ \\ \\
  \bf Facebook AI
  }

\date{}

\begin{document}
\maketitle
\begin{abstract}
Recent work has demonstrated the effectiveness of cross-lingual language model pretraining for cross-lingual understanding. In this study, we present the results of two larger multilingual masked language models, with 3.5B and 10.7B parameters. Our two new models dubbed \xlmxl and \xlmxxl outperform XLM-R by 1.8\% and 2.4\% average accuracy on XNLI. Our model also outperforms the RoBERTa-Large model on several English tasks of the GLUE benchmark by 0.3\% on average while handling 99 more languages. This suggests pretrained models with larger capacity may obtain both strong performance on high-resource languages while greatly improving low-resource languages. We make our code and models publicly available.\footnote{\url{https://github.com/pytorch/fairseq/blob/master/examples/xlmr}}
\end{abstract}

\section{Introduction}

The goal of this paper is to present a study of the impact of larger capacity models on cross-lingual language understanding (XLU).
We scale the capacity of XLM-R by almost two orders of magnitude while training on the same CC100 dataset~\cite{wenzek2019ccnet}. Our two new multilingual masked language model dubbed \xlmxl and \xlmxxl, with 3.5 and 10.7 billion parameters respectively, significantly outperform the previous XLM-R model (trained in a similar setting) on cross-lingual understanding benchmarks and obtain competitive performance with the multilingual T5 models~\cite{raffel2019exploring,xue2020mt5}. We show that they can even outperform RoBERTa-Large~\cite{roberta2019} on the GLUE benchmark~\cite{wang2018glue}.

Recent multilingual masked language models (MLM) like \mbert ~\cite{devlin2018bert} or XLM \cite{lample2019cross} improved cross-lingual language understanding by pretraining large Transformer models~\cite{transformer17} on multiple languages at once. The XLM-R model~\cite{conneau2019unsupervised} extended that approach by scaling the amount of data by two orders of magnitude, from Wikipedia to Common-Crawl and training longer, similar to RoBERTa~\cite{roberta2019}.
These models are particularly effective for low-resource languages, where both labeled and unlabeled data is scarce. They enable supervised cross-lingual transfer, where labeled data in one language can be used to solve the same task in other languages, and unsupervised cross-lingual transfer, where low-resource language self-supervised representations are improved using additional unlabeled data from higher-resource languages. Furthermore, they reduce the need for training one model per language, and allows the use of a single - potentially much larger - pretrained model that is then fine-tuned on annotated data from many languages.

The better performance of self-supervised cross-lingual models on low-resource languages comes however at the cost of lower performance on higher-resource languages~\cite{arivazhagan2019massively}. When the number of languages becomes large, \citet{conneau2019unsupervised} even observed an overall decrease of performance on all languages. It was hypothesized that when multilingual models get more capacity, they may showcase strong performance on both high-resource languages and low-resource languages. 
With only 550M parameters, the XLM-R model is now relatively small compared to new standards. Recent work scaled language models to hundreds of billions~\cite{brown2020language} or even multiple trillion parameters~\cite{fedus2021switch}, showing consistent gains in doing so. Recently, multilingual T5 showed impressive increase in performance by scaling the model capacity to tens of billions of parameters. Our study complements these findings by showing the impact of larger capacity models on the important pretraining task of \textit{multilingual} masked language modeling. We show promising results for cross-lingual understanding: \xlmxxl can both obtain a new state of the art on some cross-lingual understanding benchmarks and outperform the RoBERTa-Large model on the English GLUE benchmark~\cite{wang2018glue}. This suggests that very large-scale multilingual models may be able to benefit from the best of both worlds: obtaining strong performance on high-resource languages while still allowing for zero-shot transfer and low-resource language understanding.

\section{Pretraining and evaluation}
\label{sec:model+data}
In this section, we describe the model we use and how we scale it, as well as the data and tasks we use for pretraining and evaluation.

\subsection{Multilingual masked language models}
We use a Transformer model~\cite{transformer17} trained with the multilingual MLM objective~\cite{devlin2018bert,lample2019cross} using only monolingual data. We sample streams of text from each language and train the model to predict the masked tokens in the input. We use the same learning procedure as XLM-R.
We apply subword tokenization directly on raw text data using Sentence Piece~\cite{kudo2018sentencepiece} with a unigram language model~\cite{kudo2018subword} just like in XLM-R. We sample batches from different languages using the same sampling distribution as \citet{conneau2019unsupervised}, with $\alpha=0.3$, and without language embeddings. We use a large vocabulary size of 250K with a full softmax and train two different models: \xlmxl (L = 36, H = 2560, A = 32, 3.5B params) and \xlmxxl (L = 48, H = 4096, A = 32, 10.7B params). 
We pretrain the models on the CC100 dataset, which corresponds to 167B tokens in 100 languages. We compare our approach to previous results as well as the mT5 baselines, which were pretrained on the larger mC4 corpus of 6.4T tokens.

\insertXNLI

\subsection{Evaluation}
To evaluate our models, we use cross-lingual natural language inference and question answering for cross-lingual understanding, and the GLUE benchmark for monolingual English evaluation. 

\paragraph{Cross-lingual Natural Language Inference.}
The XNLI dataset~\cite{conneau2018xnli} comes with ground-truth dev and test sets in 15 languages, and a ground-truth English training set. The training set has been machine-translated to the remaining 14 languages, providing synthetic training data for these languages as well. We evaluate our model on cross-lingual transfer from English to other languages. We also consider two machine translation baselines: (i) \textit{translate-test}: dev and test sets are machine-translated to English and a single English model is used (ii) \textit{translate-train-all}: the English training set is machine-translated to each language and we fine-tune a multilingual model on all training sets. For the translations, we use the original data provided by the XNLI project for consistency.

\paragraph{Cross-lingual Question Answering.}
We use MLQA and XQuad benchmarks from \citet{lewis2019mlqa} and \citet{Artetxe:etal:2019}, which extend SQuAD~\cite{rajpurkar2016squad} to more languages. We report  F1 score and exact match (EM) score for cross-lingual transfer from English.

\paragraph{The English GLUE Benchmark.}
We evaluate English performance on the GLUE benchmark~\cite{wang2018glue} which gathers multiple classification tasks, such as MNLI~\cite{williams2017broad}, SST-2~\cite{socher2013recursive} or QNLI~\cite{rajpurkar2018know}. 

\subsection{Training details}
We use model parallelism based on tensor parallel~\cite{shoeybi2019megatron} for scaling models. \xlmxl uses model parallel size of 2 and \xlmxxl used 8.
Compared to previous XLM-R models, we reduce the batch size and number of updates significantly to keep the compute of the new models similar (see Table~\ref{tab:comp}). For both models, we use batch size of 2048 and train for 500,000 updates. We use pre-LayerNorm setting for both the models which was more stable during training.

For all the tasks in finetuning, we use batch size of 32 and train for 10 epochs. We do early stopping based on the average valid metrics across all languages and report test results.

\section{Analysis and Results}
\label{sec:analysis}
In this section, we present our results and compare \xlmxl and \xlmxxl performance to other methods from previous work.

\paragraph{Cross-lingual understanding results.}
On XNLI, we observe in Table~\ref{tab:xnli} that scaling the capacity from \xlml to \xlmxl leads to an average accuracy improvement of 1.4 on zero-shot cross-lingual transfer and 1.8 on multilingual fine-tuning. When scaling even further to \xlmxxl, we observe a total improvement of 2.2 on zero-shot and 2.4 on translate-train-all compared to \xlmxl, with a new state of the art on French, Vietnamese and Hindi. 
On MLQA, in Table~\ref{tab:mlqa}, we observe even larger gains for cross-lingual zero-shot transfer, where scaling from \xlml to \xlmxxl leads to improvements of 4.1 F1 and 3.9 EM scores on average. Similarly, on XQuad we observe improvements of 4.4 F1 and 5.5 scores, with new state-of-the-art results on Arabic, German, Greek and Russian (see Table~\ref{tab:xquad}).

\insertGlue

\paragraph{Comparison to monolingual English model.}
For smaller-capacity models like the Base and Large version of XLM-R, it was shown that the more languages are considered the lower the performance~\cite{conneau2019unsupervised}, in particular on high-resource languages. For instance, \xlml was outperformed by  RoBERTa\textsubscript{Large} by 1\% accuracy on average on several downstream tasks from the GLUE benchmark, as illustrated in Table\ref{tab:glue}. With larger capacity, we now observe that \xlmxxl is able to outperform RoBERTa\textsubscript{Large} by 0.3 dev points, going from 92.9 to 93.2 average accuracy, while handling 99 more languages. While a RoBERTa\textsubscript{XXL} model may outperform \xlmxxl, we believe it interesting to notice that with more capacity, a multilingual model can get strong high-resource performance while not losing its cross-lingual transfer ability for lower-resource languages.  Given the compute needed for training such large-scale models, the possibility of training a single very large model on hundreds of languages with state-of-the-art performance on high-resource languages is an encouraging result.

\insertXQuad
\insertMLQA
\insertcomparison

\paragraph{Discussion and comparison to mT5.}
Both mT5 and XLM-R models obtain strong performance on cross-lingual understanding benchmarks, as well as high performance on English benchmarks (see the score of 91.6 of mT5\textsubscript{XXL} on English XNLI).
Many hyperparameters are however different between mT5 and XLM-R models which makes difficult an apple-to-apple comparison. First, as shown in Table~\ref{tab:comp}, the mT5 models are pretrained on the much larger mC4 dataset which contains around 6.4T tokens, which is 38 times bigger than CC100 (167B tokens). While \xlml was pretrained with more updates (6T tokens), the \xlmxl and \xlmxxl models have seen less tokens (0.5T) during pretraining than their mT5 counterparts, although it also uses a bigger batch size (2048 over 1024 for mT5). Another difference is the context sequence length of 512 for XLM-R and 1024 for mT5. The mT5-XXL model also has slightly more parameters (13B over 10.7B). The larger number of updates combined with the larger dataset size may explain the larger improvement from the XL model to the XXL model in the case of mT5 (+3 average accuracy on XNLI), in which the additional capacity can exploit the large quantity of unlabeled mC4 data. We note however that the mT5\textsubscript{XL} is outperformed by \xlmxl on XNLI by 0.6\% on average, on XQuad by 1.3\% and on MLQA by 0.9\% when considering average EM score.
In comparison, gains of XLM-R from the XL to the XXL architecture are only of 0.6 on average. Another explanation may be that generative models scale better than masked language models. The difference in the nature of the pretraining dataset is particularly striking when looking at the variance of performance across languages. For example the mT5\textsubscript{XXL} outperforms \xlmxxl by 8.4 points on Swahili on XNLI zero-shot, while it only outperforms \xlmxxl by 1.4 average accuracy. These results may suggest that the CC100 dataset gets saturated with current larger-capacity models.

\section{Conclusion}
In this study, we scaled the model capacity of the XLM-R model up to 10.7B parameters and obtained stronger performance than previous XLM-R models on cross-lingual understanding benchmarks. We show that the additional capacity allows a multilingual model to outperform a the RoBERTa\textsubscript{Large} baseline on English benchmarks. Our technical study suggests that larger capacity multilingual model can obtain state-of-the-art cross-lingual understanding results while maintaining strong performance on high-resource languages. Our work provides an alternative to mT5 models, with new state-of-the-art performance on some languages, and publicly released code and models.

\bibliography{acl2021}

\begin{thebibliography}{22}
\expandafter\ifx\csname natexlab\endcsname\relax\def\natexlab#1{#1}\fi

\bibitem[{Arivazhagan et~al.(2019)Arivazhagan, Bapna, Firat, Lepikhin, Johnson,
  Krikun, Chen, Cao, Foster, Cherry et~al.}]{arivazhagan2019massively}
Naveen Arivazhagan, Ankur Bapna, Orhan Firat, Dmitry Lepikhin, Melvin Johnson,
  Maxim Krikun, Mia~Xu Chen, Yuan Cao, George Foster, Colin Cherry, et~al.
  2019.
\newblock Massively multilingual neural machine translation in the wild:
  Findings and challenges.
\newblock \emph{arXiv preprint arXiv:1907.05019}.

\bibitem[{Artetxe et~al.(2019)Artetxe, Ruder, and Yogatama}]{Artetxe:etal:2019}
Mikel Artetxe, Sebastian Ruder, and Dani Yogatama. 2019.
\newblock On the cross-lingual transferability of monolingual representations.
\newblock \emph{arXiv preprint arXiv:1910.11856}.

\bibitem[{Brown et~al.(2020)Brown, Mann, Ryder, Subbiah, Kaplan, Dhariwal,
  Neelakantan, Shyam, Sastry, Askell et~al.}]{brown2020language}
Tom~B Brown, Benjamin Mann, Nick Ryder, Melanie Subbiah, Jared Kaplan, Prafulla
  Dhariwal, Arvind Neelakantan, Pranav Shyam, Girish Sastry, Amanda Askell,
  et~al. 2020.
\newblock Language models are few-shot learners.
\newblock \emph{Proc. of NeurIPS}.

\bibitem[{Conneau et~al.(2019)Conneau, Khandelwal, Goyal, Chaudhary, Wenzek,
  Guzm{\'a}n, Grave, Ott, Zettlemoyer, and Stoyanov}]{conneau2019unsupervised}
Alexis Conneau, Kartikay Khandelwal, Naman Goyal, Vishrav Chaudhary, Guillaume
  Wenzek, Francisco Guzm{\'a}n, Edouard Grave, Myle Ott, Luke Zettlemoyer, and
  Veselin Stoyanov. 2019.
\newblock Unsupervised cross-lingual representation learning at scale.
\newblock \emph{arXiv preprint arXiv:1911.02116}.

\bibitem[{Conneau et~al.(2018)Conneau, Rinott, Lample, Williams, Bowman,
  Schwenk, and Stoyanov}]{conneau2018xnli}
Alexis Conneau, Ruty Rinott, Guillaume Lample, Adina Williams, Samuel~R.
  Bowman, Holger Schwenk, and Veselin Stoyanov. 2018.
\newblock Xnli: Evaluating cross-lingual sentence representations.
\newblock In \emph{EMNLP}. Association for Computational Linguistics.

\bibitem[{Devlin et~al.(2018)Devlin, Chang, Lee, and
  Toutanova}]{devlin2018bert}
Jacob Devlin, Ming-Wei Chang, Kenton Lee, and Kristina Toutanova. 2018.
\newblock Bert: Pre-training of deep bidirectional transformers for language
  understanding.
\newblock \emph{NAACL}.

\bibitem[{Fedus et~al.(2021)Fedus, Zoph, and Shazeer}]{fedus2021switch}
William Fedus, Barret Zoph, and Noam Shazeer. 2021.
\newblock Switch transformers: Scaling to trillion parameter models with simple
  and efficient sparsity.
\newblock \emph{arXiv preprint arXiv:2101.03961}.

\bibitem[{Kudo(2018)}]{kudo2018subword}
Taku Kudo. 2018.
\newblock Subword regularization: Improving neural network translation models
  with multiple subword candidates.
\newblock In \emph{ACL}, pages 66--75.

\bibitem[{Kudo and Richardson(2018)}]{kudo2018sentencepiece}
Taku Kudo and John Richardson. 2018.
\newblock Sentencepiece: A simple and language independent subword tokenizer
  and detokenizer for neural text processing.
\newblock \emph{EMNLP}.

\bibitem[{Lample and Conneau(2019)}]{lample2019cross}
Guillaume Lample and Alexis Conneau. 2019.
\newblock Cross-lingual language model pretraining.
\newblock \emph{NeurIPS}.

\bibitem[{Lewis et~al.(2019)Lewis, O\u{g}uz, Rinott, Riedel, and
  Schwenk}]{lewis2019mlqa}
Patrick Lewis, Barlas O\u{g}uz, Ruty Rinott, Sebastian Riedel, and Holger
  Schwenk. 2019.
\newblock Mlqa: Evaluating cross-lingual extractive question answering.
\newblock \emph{arXiv preprint arXiv:1910.07475}.

\bibitem[{Liu et~al.(2019)Liu, Ott, Goyal, Du, Joshi, Chen, Levy, Lewis,
  Zettlemoyer, and Stoyanov}]{roberta2019}
Yinhan Liu, Myle Ott, Naman Goyal, Jingfei Du, Mandar Joshi, Danqi Chen, Omer
  Levy, Mike Lewis, Luke Zettlemoyer, and Veselin Stoyanov. 2019.
\newblock Roberta: {A} robustly optimized {BERT} pretraining approach.
\newblock \emph{arXiv preprint arXiv:1907.11692}.

\bibitem[{Raffel et~al.(2019)Raffel, Shazeer, Roberts, Lee, Narang, Matena,
  Zhou, Li, and Liu}]{raffel2019exploring}
Colin Raffel, Noam Shazeer, Adam Roberts, Katherine Lee, Sharan Narang, Michael
  Matena, Yanqi Zhou, Wei Li, and Peter~J. Liu. 2019.
\newblock Exploring the limits of transfer learning with a unified text-to-text
  transformer.
\newblock \emph{arXiv preprint arXiv:1910.10683}.

\bibitem[{Rajpurkar et~al.(2018)Rajpurkar, Jia, and Liang}]{rajpurkar2018know}
Pranav Rajpurkar, Robin Jia, and Percy Liang. 2018.
\newblock Know what you don't know: Unanswerable questions for squad.
\newblock \emph{ACL}.

\bibitem[{Rajpurkar et~al.(2016)Rajpurkar, Zhang, Lopyrev, and
  Liang}]{rajpurkar2016squad}
Pranav Rajpurkar, Jian Zhang, Konstantin Lopyrev, and Percy Liang. 2016.
\newblock Squad: 100,000+ questions for machine comprehension of text.
\newblock \emph{arXiv preprint arXiv:1606.05250}.

\bibitem[{Shoeybi et~al.(2019)Shoeybi, Patwary, Puri, LeGresley, Casper, and
  Catanzaro}]{shoeybi2019megatron}
Mohammad Shoeybi, Mostofa Patwary, Raul Puri, Patrick LeGresley, Jared Casper,
  and Bryan Catanzaro. 2019.
\newblock Megatron-lm: Training multi-billion parameter language models using
  model parallelism.
\newblock \emph{arXiv preprint arXiv:1909.08053}.

\bibitem[{Socher et~al.(2013)Socher, Perelygin, Wu, Chuang, Manning, Ng, and
  Potts}]{socher2013recursive}
Richard Socher, Alex Perelygin, Jean Wu, Jason Chuang, Christopher~D Manning,
  Andrew Ng, and Christopher Potts. 2013.
\newblock Recursive deep models for semantic compositionality over a sentiment
  treebank.
\newblock In \emph{EMNLP}, pages 1631--1642.

\bibitem[{Vaswani et~al.(2017)Vaswani, Shazeer, Parmar, Uszkoreit, Jones,
  Gomez, Kaiser, and Polosukhin}]{transformer17}
Ashish Vaswani, Noam Shazeer, Niki Parmar, Jakob Uszkoreit, Llion Jones,
  Aidan~N. Gomez, Lukasz Kaiser, and Illia Polosukhin. 2017.
\newblock Attention is all you need.
\newblock In \emph{Advances in Neural Information Processing Systems}, pages
  6000--6010.

\bibitem[{Wang et~al.(2018)Wang, Singh, Michael, Hill, Levy, and
  Bowman}]{wang2018glue}
Alex Wang, Amanpreet Singh, Julian Michael, Felix Hill, Omer Levy, and Samuel~R
  Bowman. 2018.
\newblock Glue: A multi-task benchmark and analysis platform for natural
  language understanding.
\newblock \emph{arXiv preprint arXiv:1804.07461}.

\bibitem[{Wenzek et~al.(2019)Wenzek, Lachaux, Conneau, Chaudhary, Guzman,
  Joulin, and Grave}]{wenzek2019ccnet}
Guillaume Wenzek, Marie-Anne Lachaux, Alexis Conneau, Vishrav Chaudhary,
  Francisco Guzman, Armand Joulin, and Edouard Grave. 2019.
\newblock Ccnet: Extracting high quality monolingual datasets from web crawl
  data.
\newblock \emph{arXiv preprint arXiv:1911.00359}.

\bibitem[{Williams et~al.(2017)Williams, Nangia, and
  Bowman}]{williams2017broad}
Adina Williams, Nikita Nangia, and Samuel~R Bowman. 2017.
\newblock A broad-coverage challenge corpus for sentence understanding through
  inference.
\newblock \emph{Proceedings of the 2nd Workshop on Evaluating Vector-Space
  Representations for NLP}.

\bibitem[{Xue et~al.(2020)Xue, Constant, Roberts, Kale, Al-Rfou, Siddhant,
  Barua, and Raffel}]{xue2020mt5}
Linting Xue, Noah Constant, Adam Roberts, Mihir Kale, Rami Al-Rfou, Aditya
  Siddhant, Aditya Barua, and Colin Raffel. 2020.
\newblock mt5: A massively multilingual pre-trained text-to-text transformer.
\newblock \emph{arXiv preprint arXiv:2010.11934}.

\end{thebibliography}
\bibliographystyle{acl_natbib}

\end{document}